\documentclass[letterpaper, 10 pt, conference]{ieeeconf}  

\IEEEoverridecommandlockouts                              

\usepackage[noadjust]{cite}
\usepackage{amssymb}
\usepackage{amsmath}
\usepackage{graphics}
\usepackage{epsfig}
\usepackage{breqn}
\usepackage[T1]{fontenc}
\usepackage{booktabs,multirow,array}
\usepackage[pagebackref=true,colorlinks,citecolor=green]{hyperref}
\hypersetup{%
  colorlinks = True,
  linkcolor  = blue
}

\usepackage[justification=centering]{caption}
\let\labelindent\relax
\usepackage[inline]{enumitem}  

\usepackage{pgfplots}

\usepackage[normalem]{ ulem }
\usepackage{soul}

\title{\LARGE \bf
PreTR: Spatio-Temporal Non-Autoregressive Trajectory Prediction Transformer
}

\author{Lina Achaji$^{1,2}$, Thierno Barry$^{1}$, Thibault Fouqueray$^{1}$, Julien Moreau$^{1}$, Francois Aioun$^{1}$, and Francois Charpillet$^{2}$
\thanks{$^{1}$Stellantis Group, Technical center of Velizy 78140, France
{\tt\small lina.achaji@stellantis.com}
{\tt\small thibault.fouqueray@stellantis.com}
{\tt\small julien.moreau@stellantis.com}
{\tt\small francois.aioun@stellantis.com}}
\thanks{$^{2}$Université de Lorraine, CNRS, Inria, LORIA, Nancy 54000, France
        {\tt\small francois.charpillet@inria.fr}}%
}

\begin{document}

\maketitle
\thispagestyle{empty}
\pagestyle{empty}

\begin{abstract}

Nowadays, our mobility systems are evolving into the era of intelligent vehicles that aim to improve road safety. Due to their vulnerability, pedestrians are the users who will benefit the most from these developments. However, predicting their trajectory is one of the most challenging concerns. Indeed, accurate prediction requires a good understanding of multi-agent interactions that can be complex. Learning the underlying spatial and temporal patterns caused by these interactions is even more of a competitive and open problem that many researchers are tackling. In this paper, we introduce a model called PRediction Transformer (PReTR) that extracts features from the multi-agent scenes by employing a factorized spatio-temporal attention module. It shows less computational needs than previously studied models with empirically better results. 
Besides, previous works in motion prediction suffer from the exposure bias problem caused by generating future sequences conditioned on model prediction samples rather than ground-truth samples. In order to go beyond the proposed solutions, we leverage encoder-decoder Transformer networks for parallel decoding a set of learned object queries. This non-autoregressive solution avoids the need for iterative conditioning and arguably decreases training and testing computational time. We evaluate our model on the ETH/UCY datasets, a publicly available benchmark for pedestrian trajectory prediction. Finally, we justify our usage of the parallel decoding technique by showing that the trajectory prediction task can be better solved as a non-autoregressive task.

\end{abstract}


\section{Introduction}

The introduction of intelligent driving solutions has the potential of improving road safety, especially in urban areas. For instance, vulnerable road users (i.e., cyclists and pedestrians) will be the most affected by improved road safety, as they are the least protected road users \cite{yannis2020vulnerable}. In fact, one crucial challenge in intelligent urban solutions is to predict the future behavior of these road users. The prediction block focuses on understanding the latent intention of the agents and generating their possible future trajectories. These intentions can be very complex due to uncertainties related to single human motion and human-human interactions. Building a model capable of learning multi-agent interaction requires temporal modeling of an agent's past states as well as a social understanding of other agents' influence on its future trajectory. In fact, multi-agent interaction modeling has been previously addressed using social spatio-temporal LSTM-based models \cite{alahi2016social, gupta2018social, xue2018ss}. However, LSTM models are not competent enough to process long-term dependencies in sequences due to the vanishing gradient problem \cite{hochreiter1998vanishing}. To avoid this problem, \cite{MTN_trajectory, yuan2021agent} proposed Transformer-based models \cite{vaswani2017attention} for multi-agent trajectory forecasting. Even though Transformer models were not originally designed to handle social and temporal dependencies, it becomes feasible to apply them in multi-agent scenarios by adapting the architecture of the attention module. The authors in \cite{yuan2021agent} suggested using an attention module that receives the flattened input states for multiple pedestrians at different time steps in a merged fashion. By choosing this merged-attention design, they argued that their model can capture the dependency between multiple agents at distinct time steps. Nevertheless, in our work, we show that even if this model can be slightly beneficial in some cases, it comes with a cost on training and inference computational time and memory. In this paper, we propose to use a divided spatio-temporal architecture \cite{bertasius2021space} originally applied for video understanding. It factorizes the attention module on the temporal and spatial axis separately instead of processing them jointly. This model shows an advantage in computational time compared to previously proposed solutions.

In fact, Transformer models have outperformed LSTM models in different areas, such as in \cite{radford2019language, yang2019xlnet, aksan2020attention}. Their main advantages lie in their outstanding performance due to attention mechanism and parallelization during training. The parallelization engenders computational time excellency using the teacher-forcing \cite{lamb2016professor} techniques.
However, the prediction remains auto-regressive during inference (i.e., generates time-steps iteratively), resulting in a slow inference problem. In addition, training the model in teacher forcing mode and testing it in an auto-regressive fashion has been shown to cause the exposure bias problem \cite{ranzato2015sequence, schmidt2019generalization, chiang2021relating}. It is defined by the train-test discrepancy causing the model to accumulate error over time when exposed to its own instead of ground-truth previous step predictions. To address this problem, \cite{yuan2021agent} trained the Transformer model autoregressively instead of using the teacher forcing approach. However, this prevents the Transformer architecture from using its parallelization power, leading to a large increase in training time. 

Recently, \cite{carion2020end} has proposed a detection Transformer model called DETR used for object detection. DETR applies attention modules on a set of learned inputs called queries to directly detect the class and bounding box position of objects inside the image in a parallel manner. This model is promising because it can retain the parallelization property in Transformers. It has recently been used for object tracking using segmentation \cite{sun2020transtrack, meinhardt2021trackformer}. Although tracking is a relatively close problem to prediction, they only used the parallel decoding property to detect objects in the scene. Then, an autoregressive approach was applied to track the detected objects over time. In this paper, we leverage the DETR-like parallel decoding strategy for directly predicting spatio-temporal interaction aware trajectories without using the auto-regressive training. This model, called Prediction Transformer (PreTR), will show an advantage in both train and test time while preserving evaluation metrics performances in trajectory prediction applications. 

Finally, we raise the question of the nature of the trajectory prediction problem. We find that it can be considered a non-autoregressive task. This finding confirms our parallel decoding empirical results and changes the conventional view to the trajectory prediction problem. 

In summary, our contributions are three-fold:
\begin{itemize}
    \item We addressed the social-temporal interaction problem through a factorized spatio-temporal attention module.
    \item We leveraged a parallel-decoding strategy that reduces computational time and eliminates the exposure bias problem in autoregressive models.
    \item We worked on a self-supervised spatio-temporal model that was able to show the non-autoregressive nature of the trajectory prediction task and justify our outstanding parallel-decoding results.
\end{itemize}

\section{Related Work}
\subsection{Pedestrian Trajectory Prediction}
Pedestrian trajectory prediction methods can be categorized based on different model factors such as context and environment awareness, social and agent-interaction awareness, multi-modal predictions, and auto-regressive or parallel predictions. Social and agent-interaction awareness has been studied in multiple works. In fact, the Social-LSTM \cite{alahi2016social} model used a deterministic LSTM-based method coupled with a social-pooling strategy to connect neighboring pedestrians during learning. The authors in \cite{gupta2018social} used the same interaction model in an adversarial training scenario to predict multi-modal plausible social-aware trajectories. Sophie \cite{sadeghian2019sophie} added physical- and social-attention modules to an LSTM-based GAN model in a context-aware approach. Recently, AgentFormer \cite{yuan2021agent} has used the Transformer model for joint spatial and temporal modeling that shows an advantage over the individual-aware Transformer model used in \cite{giuliari2021transformer}. AgentFormer was coupled with a conditional variational auto-encoder (CVAE) \cite{sohn2015learning} to account for trajectory multi-modality. These before-mentioned models use auto-regressive decoding during prediction. Although non-autoregressive prediction is an active research area in sequence generation \cite{ren2020study} and machine translation \cite{gu2017non} applications, we note a modest contribution to the trajectory prediction problem. For instance, the authors in \cite{xue2020take} proposed an LSTM-based non-autoregressive pedestrian trajectory predictor employing context generators to compensate for the loss caused by the non-sequential dependencies during prediction. 

\subsection{Parallel Decoding Models}
Non-autoregressive techniques are recently studied to decrease inference computational time compared to auto-regressive approaches. The model in \cite{gu2017non} generates all the tokens in the target sequence in a parallel fashion without taking into account the token-to-token dependency. They introduced the fertility prediction concept to address the multi-modality problem in language translation. The authors in \cite{zhou2019understanding} justified the difficulty of non-autoregressive training in some applications. Alternatively, \cite{ghazvininejad2019mask, chen2020non} leveraged the usage of conditional masking for non-autoregressive generation in machine translation and speech recognition. Conditioned on the input sequence and a partially masked target sequence, the models are trained to predict the unmasked subset of the target tokens. Recently, a model called DETR has been proposed in \cite{carion2020end} for object detection. Its decoder takes a set of learned positional embeddings, called object queries, and independently predicts their corresponding box coordinates using a parallel decoding technique. This technique was feasible due to the parallelization capabilities of the Transformer model. POTR model \cite{martinez2021pose} exploited the object queries for skeleton motion prediction. They outperformed the autoregressive solution in long-term prediction horizons and computational time.

\section{Method}

We formulate the trajectory prediction problem as a scene prediction of $N$ observed pedestrians. The scenes are dynamic with a variable number of pedestrians at different time steps. An observation trajectory can be represented by a sequence $X = \{X_1, . . . , X_{T\_obs} \}$ where $X_t = \{a_t^1,...,a_t^N\}$, $t \in \{1, ..., T\_obs\}$ denotes an agent or pedestrian at time step $t$ with states $a_t^n \in{\mathbb{R}^{d_{inp}}}$, where $d_{inp}$ is the input dimension. The input states include the $2D$ position and velocity of each pedestrian ($d\_inp = 4$).
We receive the inputs of all observed pedestrians and simultaneously predict their corresponding future states $Y = \{Y_{T\_obs+1}, . . . , Y_{T\_pred}\}$, where $t \in \{1, ..., T\_pred\}$ and $Y_t \in{\mathbb{R}^{N*d_{out}}}$. $d_{out}$ denotes the output dimension that includes the $2D$ position coordinates for each predicted pedestrian ($d\_out = 2$). Our aim is to learn a conditional generative model $p(Y|X)$.

\begin{figure*}
  \includegraphics[scale=0.4]{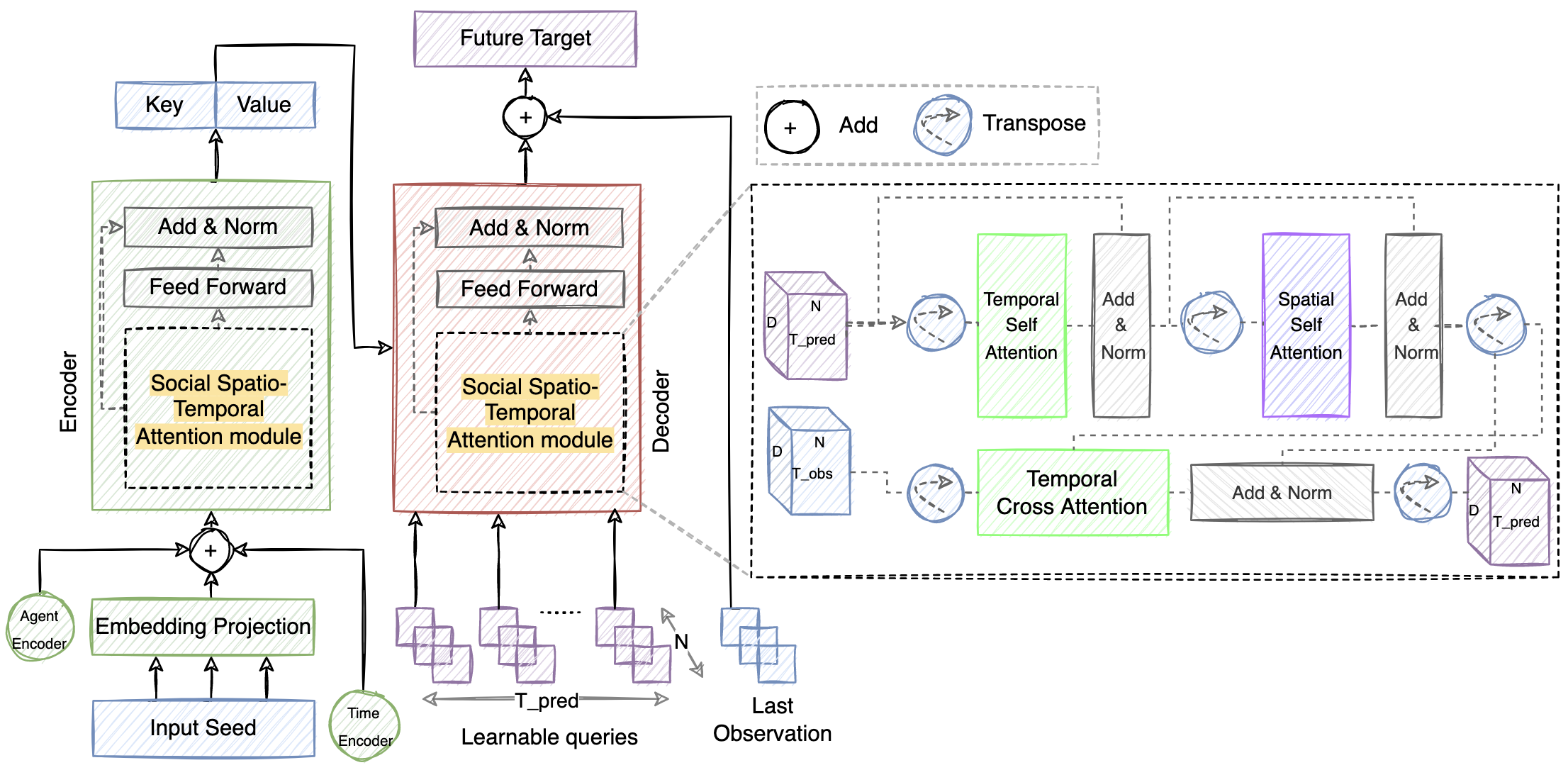}
\centering
\caption{Overview of the PreTR model. Our model based on the encoder-decoder formalism in \cite{vaswani2017attention} takes the embedded source input seed sequence into the model encoder. The decoder takes as input a $3-$dimensional set of learned queries that correspond to the prediction length ($T_{pred}$), the number of agents $N$, and the embedding dimension $D$. The decoder predicts the residual between the last observed input (in blue) and the final prediction for each time step of each agent.  }
\label{fig:archi}
\end{figure*}

\subsection{PreTR model}
Our model uses the Transformer encoder-decoder paradigm used in \cite{vaswani2017attention} with multiple variants added.  First, the classic Transformer is designed for one-dimensional temporal data. Our model extends the original attention module to account for spatial and temporal multi-agent dependencies. Second, \cite{vaswani2017attention} adopts an autoregressive strategy in the decoding procedure. It follows the teacher forcing technique during training by taking the original target sequence as the input to the decoder. However, since we do not have access to the ground truth during inference, it applies fully autoregressive decoding where the model conditions its prediction to generate the following steps. 
Instead, our model unifies the decoding process in the training and inference stages by using a non-autoregressive procedure to predict the target at all time steps independently.
\subsubsection{Input Embedding}
The model takes as input a 3-dimensional vector $X \in \mathbb{R}^{{T\_obs}*N*D\_inp}$ and goes into the encoder embedding module.

\begin{equation}
  X = \begin{bmatrix}
a_1^1  & a_1^2 & \ldots & a_1^N\\
a_2^1  & a_1^2 & \ldots & a_2^N\\
\vdots  & \vdots & \ddots & \vdots\\
a_{T\_obs}^1 & a_{T\_obs}^2 & \ldots & a_{T\_obs}^N\\
\end{bmatrix}
\end{equation}

The embedding module projects the X input into a $D-$dimensional space using a linear layer:
$E = WX + b$, where $E \in \mathbb{R}^{T*N*D}$. The $W$ and $b$ are the corresponding weights and biases in the embedding layer. 
\subsubsection{Time Encoding} The attention module evaluates the input as an unordered set of tokens. However, temporal ordering is essential for sequential prediction applications. Thus, we apply a temporal encoder on top of the embedding layer to introduce explicit time ordering. We define the time encoder by a parametric layer, randomly initialized, where the number of parameters corresponds to the observation sequence length. In this manner, we learn the time ordering encoding end-to-end while training the prediction model, $\Theta_T \in \mathbb{R}^{T\_obs}$. We addition the time encoding to the projected embedding vector $E$.
\subsubsection{Agent Encoding} 
The loss of sequence information is also a problem for preserving the identity of agents. Although the sequential order of agents is not natural, the model must still assign all states of a particular agent at different time steps and differentiate its index from other agents. As for the temporal encoding, we represent the agent encoding by a parametric layer where the number of parameters is set to the maximum number of agents, $\Theta_A \in \mathbb{R}^{N}$. To prevent the model from learning a relationship between agents based on their index in the scene, we apply an agent permutation technique on the input and target of the model. As a result, the agent encoder will learn to maintain agent identity without memorizing a specific order-related behavior for the agents.

\subsubsection{Social spatio-temporal Attention Module}\label{axial} The core of our Transformer network is a divided attention module that operates on both spatial and temporal domains. We factorize the attention into two consecutive classical attention modules (Fig. \ref{fig:archi}). The first module computes temporal self-attention ($TAttn$), by transposing the temporal and spatial axes of the input embedded vector. In this way, we consider each agent independently over all time steps. Then, the corresponding output is passed to the spatial self-attention ($SAttn$) module, which attends to all agents for each time step. Each self-attention block is an operation over the query, key, and value vectors:

\begin{dmath}
   Attention(Q,K,V) = softmax( QK^T / \sqrt D + M) V = AV 
\end{dmath}
The queries ($Q$), keys ($K$), and values ($V$) are the parametric linear projections of the input embedding vector. In the temporal case, the attention weights $A$ denote the relative score given to each time step compared to the other time steps for each agent $n$. For the spatial attention block, $A$ represents the associated relative score of each agent $i$ regarding another agent $j$ at each time $t$. The attention module can be extended using the multi-head attention mechanism ($MHA$) described in \cite{vaswani2017attention}.
The mask $M$ was designed by \cite{vaswani2017attention} to preserve causality in autoregressive predictions. However, since our decoding strategy follows a non-autoregressive approach, it is not needed for this purpose anymore. Nevertheless, we used masking on the attention maps to account for missing time-steps or agents in the seed trajectories. Finally, we add the output of the each spatio-temporal attention to the original input embedded vector followed by a layer normalization \cite{vaswani2017attention}.

The self-attention complexity is proportional to $O(n^2 \cdot d)$, where $n$ is the sequence length and $d$ is the model embedding dimension. Our divided attention module has a time-computational advantage over the merged attention module described in \cite{yuan2021agent}. For instance, the temporal and spatial attentions have a complexity of $O(T^2 \cdot N \cdot D)$ and  $O(N^2 \cdot T  \cdot D)$ respectively. Thus, the divided attention has a total complexity of $O((T \cdot N)(T+N) D)$. Otherwise, the merged attention in \cite{yuan2021agent} flattens the whole sequence resulting in an attention module with complexity proportional to $O(T^2 \cdot N^2 \cdot D)$.

\subsubsection{Encoder}\label{encoder}
The encoder block takes the $D-$dimensional embedded vector $E$ as input. It is composed of multiple encoder layers. Each encoder layer contains the previously mentioned spatio-temporal attention block alongside a point-wise feed-forward ($Pff$) followed by an additional normalization layer ($LN$). The Encoder layer can be summarized using the following equations:
\begin{equation}
E_t = LN(E + TAttn(E,E,E))
\end{equation}
\begin{equation}
E_{ts} = LN(E_t + SAttn(E_t, E_t, E_t))
\end{equation}
\begin{equation}
E = LN( Pff( E_{ts} ) + E_{ts})
\end{equation}

\subsubsection{Decoder}\label{decoder}
The decoder block is in charge of generating the prediction sequence in a parallel fashion. It receives as input a set of learned object queries ($QL$) of dimension $D$ conforming to the encoder's embedding dimension. They are randomly initialized and then end-to-end optimized during model training.
The temporal dimension of these queries is fixed to the desired output length. The spatial parameters size matches the number of agents in the seed sequence. The embedding of the input is unnecessary since the queries are $D-$dimensional by design. However, we have added the temporal and the agent encoding similar to the encoder encoding. 
The decoder is composed of several decoding layers. Each of these contains the spatio-temporal attention block alongside the point-wise feed-forward and layer normalization layers. The decoder spatio-temporal block includes an additional temporal attention layer responsible for cross-attention between the decoder's self-attention output and the encoder's memory vector ($E_m$). We added only temporal cross-attention instead of fully spatio-temporal cross-attention since the length of the temporal prediction is different from the seed sequence length. The decoder layer can be summarised using the following equations:

\begin{equation}
E_t = LN(QL + TAttn(QL,QL,QL))
\end{equation}
\begin{equation}
E_{ts} = LN(E_t + SAttn(E_t, E_t, E_t))
\end{equation}
\begin{equation}
E_{cts} = LN(E_{ts} + TAttn(E_{ts},E_m,E_m))
\end{equation}
\begin{equation}
E = LN( Pff( E_{cts} ) + E_{cts})
\end{equation}

At the final stage of the decoder, we add an inverse embedding layer responsible for trajectory generation of the 2-dimensional sequence coordinates. We add the last observed seed input to the decoder output. In this way, the decoder predicts the variation between the seed and the generated steps for each pedestrian. 

The parallel decoder block has lower computational complexity than autoregressive decoders. In fact, for autoregressive decoders, we can take advantage of the parallelization power of Transformers by applying the teacher forcing strategy. However, this leads to the problem of exposure bias arising from the fact that the model will not be able to compensate for its prediction errors. As a possible solution, the model can be trained and tested in an autoregressive strategy with the cost of a large increase in training time. In both cases, the autoregressive approach is costly during inference, requiring an iterative pass over the model predictions at each step. In contrast, a parallel decoding strategy can be applied uniformly during training and inference, eliminating the source of an exposure bias problem. Moreover, it has the same complexity as the teacher's forcing strategy, allowing parallelization to be introduced during the inference stage. For instance, the parallel decoder has a complexity of $O(T^2D)$, while the autoregressive has a higher complexity of $O((T)(T+1)(2T+1)D)$ during inference.

\subsection{Model Training and Inference}
We train the PreTR model with an end-to-end fashion to predict the whole target sequence in a single forward pass. We use the mean-squared error distance between the ground truth and the generated target:

\begin{equation}
    L = \frac{1}{T N} \; \sum_{t=T_{obs} + 1}^{T_{pred}} \; \sum_{n=1}^{N}(\hat{a}_t^n-a_t^n)^2
\end{equation}

where $ T = T_{pred} - T_{obs}$ corresponds to the prediction length, and $N$ is the number of observed agents. Since our model is non-autoregressive, the inference and training stages are identical without further changes.

\begin{table*}[t]
\caption{Baseline comparison following the single-trajectory deterministic approach on the ETH/UCY Datasets. Some of the results are taken from \cite{giuliari2021transformer, salzmann2020trajectron}. * is evaluated by us since the deterministic result was not published in \cite{yuan2021agent}.}
    \resizebox{\textwidth}{!}{%
        \begin{tabular}{l  c c c c c c c| c}
          \toprule
          \multirow{2}{8em}{Method} & \multirow{2}{7em}{Social Modeling} & \multirow{2}{8em}{Decoding method} & \multicolumn{6}{c}{ADE/FDE [meter]} \\
          & & &  ETH & Hotel & Univ & Zara1 & Zara2 & Average \\
          \midrule
          LSTM \cite{gupta2018social} & Non & Auto Regressive & 1.09/2.94 & 0.86/1.91 & 0.61/1.31 & 0.41/0.88 & 0.52/1.11 & 0.70/1.52\\

          S-GAN-ind \cite{gupta2018social} & Non & Auto Regressive & 1.13/2.21 & 1.01/2.18 & 0.60/1.28 & 0.42/0.91 & 0.52/1.11 & 0.74/1.54\\

        Transformers-TF \cite{giuliari2021transformer} & Non & Auto Regressive & 1.03/2.10 & 0.36/0.71 & 0.53/1.32 & 0.44/1.00 & 0.34/0.76 & 0.54/1.17\\
        \midrule
        \midrule
        
          Social LSTM \cite{alahi2016social} & Social Pooling & Auto Regressive & 1.09/2.35 & 0.79/1.76 & 0.67/1.40 & 0.47/1.00 & 0.56/1.17 & 0.72/1.54\\

          Social Attention \cite{kosaraju2019socialattention} & Social Pooling & Auto Regressive & 0.39/3.74 & 0.29/2.64 & 0.20/0.52 & 0.30/2.13 & 0.33/3.92 & 0.30/2.59\\

          AgentFormer\footnote[2]{}* \cite{yuan2021agent} & Merged Attention & Auto Regressive & 1.02/2.02 & 0.35/0.67 & 0.64/1.31 & 0.46/0.98 & 0.37/0.81 & 0.57/1.16 \\
          \midrule
          Ours & Divided Attention & Auto Regressive & 0.9/1.62 & 0.36/0.61 & 0.9/1.56 & 0.47/0.89 & 0.42/0.78 & 0.61/1.07 \\
          Ours (PreTR) & Divided Attention & Parallel & \textbf{0.82/1.55} & \textbf{0.3/0.56} & \textbf{0.62/1.23} & \textbf{0.42/0.89} & \textbf{0.35/0.73} & \textbf{0.50/0.99} \\
          \bottomrule

        \end{tabular}
    }
    \label{table:deterministic_ethucy}
\end{table*}

\section{Experiments on the PreTR model}
\subsection{Datasets and evaluation protocols}
We trained and evaluated the PreTR model on the ETH/UCY \cite{pellegrini2009you, lerner2007crowds} benchmark datasets for pedestrian multi-agent trajectory prediction.
\subsubsection{ETH/UCY datasets} include five different datasets at four unique locations with a total number of 1536 pedestrians. The locations are captured with a bird's eye view perspective and annotated at 2.5 FPS (frame per second). For the evaluation protocol, we adopt the leave-one-out cross-validation strategy as in \cite{salzmann2020trajectron}. It consists of training the model on four of the datasets and testing on the one left dataset. We trained our model to generate 12 (4.8 sec) future timesteps in each agent's trajectory using a seed sequence of 8 (3.2 sec) observed timesteps.

\subsection{Evaluation Metrics}
Since our model predicts a single trajectory for each pedestrian, we report the following uni-modal evaluation metrics such as in \cite{salzmann2020trajectron, giuliari2021transformer, kothari2021human}:

\textit{Average displacement error (ADE):} the average euclidean distance between the predicted trajectories and the ground truth over all time steps.

\textit{Final displacement error (FDE):} the final euclidean distance between the predicted trajectories and the ground truth at the last time step.

\subsection{Implementation details}
For each pedestrian in the datasets, we concatenated its two-dimensional speed with its position to obtain a 4-dimensional model input. As mentioned earlier, the scenes in the datasets are dynamic and contain a varying number of pedestrians at different time steps. Furthermore, in order to use mini-batch learning, all scenes should have the same number of agents. Thus, during the training stage, we restricted the maximum number of agents to $n_{th} = 20$ and padded the scenes with fewer agents. Then, we applied a padding mask on the attention weights of the model to handle missing and padded observations. During the model evaluation, we use an additional mask on the loss and metric functions to only consider presented ground truth time steps. The different datasets in the ETH/UCY benchmark are not all captured using the same standards. Thus, we normalized each of the five datasets to contain trajectory values between $0$ and $1$ according to its minimum and maximum coordinates values. We applied data augmentation by rotation on the training and validation sets. To prevent agent index-based behavior memorization, we shuffled the indices of the agents in each scene during training. For all experiments, we use one or two encoder-decoder model layer (depending on the dataset) and $8$ multi-heads in the attention layers. We set the embedding dimension and $Pff$ layer hidden size to $256$ and $512$. During optimization, we use the Adam optimizer \cite{kingma2014adam} with the warmup steps strategy \cite{vaswani2017attention}. We set the warmup steps to $2500$.

\subsection{Results}\label{results}

We compared our model against the state-of-the-art baselines on the ETH/UCY datasets that employ the same environmental approach as ours (only information about 2-dimensional coordinates without using environmental maps or images) on the single-trajectory deterministic prediction. First of all, as we see in Table 2, the PreTR model outperforms the LSTM based models on the FDE score (-0.51). Furthermore, PreTR surpasses the recent AgentFormer\footnote[2]{We evaluated AgentFormer on the single trajectory prediction approach by extending the originally published code to the deterministic case: \url{https://github.com/Khrylx/AgentFormer}}* \cite{yuan2021agent} model that uses agent-aware merged attention in their architecture in terms of ADE (-0.06) and FDE (-0.15) scores. Our model's outstanding results are coming from both the attention module that can handle long-term dependencies better than LSTM-based social modules, along with the advantage of using parallel decoding for trajectory generation. We can see this finding by evaluating the divided attention module in the autoregressive decoding framework. This model beats the merged-attention module in the FDE score with an advantage on long-term horizon predictions. However, the parallel decoder approach remains better than the auto-regressive approach on short-term and long-term prediction (-0.1 / -0.06). In fact, instead of generating the next steps recurrently, our model is more adapted to predict a global latent representation using the seed sequence leading to better long-term planning. We will later reflect on the parallel decoding effect in the section \ref{comma}.
We showed in Table \ref{table:inference} the advantages of both divided attention and parallel decoding in terms of inference time speed-up. If we consider autoregressive decoding, the divided attention module reaches twice the speed of the merged attention solution. In addition, divided attention in the parallel decoding strategy is up to 13 times faster than in the autoregressive decoding approach. We calculate these results on a prediction horizon of 4.8 seconds. Indeed, the acceleration will increase linearly with larger prediction horizons. Furthermore, the parallel decoder also shows the advantage over autoregressive training per batch time. The model in \cite{giuliari2021transformer} uses parallelization since it applies the teacher forcing strategy. We should note that the number of epochs needed for convergence is similar for the parallel and autoregressive approaches. All reported inference times are estimated on an Nvidia-GPU 1080Ti. 

\begin{table}[t]
\caption{Transformer-based models speed comparison on the inference (I), and training (T) [per batch] stage.}
\begin{center}
        \begin{tabular}{c | c c  c | c c}
          \toprule
          & \multicolumn{3}{c|}{Autoregressive} & \multicolumn{2}{c}{Parallel}\\
    & Temporal & Divided & Merged & Temporal & Divided\\
    & \cite{giuliari2021transformer} & (Ours) & \cite{yuan2021agent} & (Ours) & (Ours) \\
          \midrule
          \midrule
          Inference & \multirow{2}{*}{8.63} & \multirow{2}{*}{10.92} & \multirow{2}{*}{20.9} & \multirow{2}{*}{1.1} & \multirow{2}{*}{1.53}  \\
          Time [ms] & & & & &  \\
          \midrule
          I/T & \multirow{2}{*}{2.5/19 $\times$} & \multirow{2}{*}{2/2 $\times$ } & \multirow{2}{*}{1/1 $\times$} & \multirow{2}{*}{19/19 $\times$}& \multirow{2}{*}{11/11 $\times$}\\
          Speed-Up & & & & &\\
          \bottomrule
        \end{tabular}
\end{center}    
    \label{table:inference}
\end{table}

\begin{figure}[b]
  \includegraphics[scale=0.35]{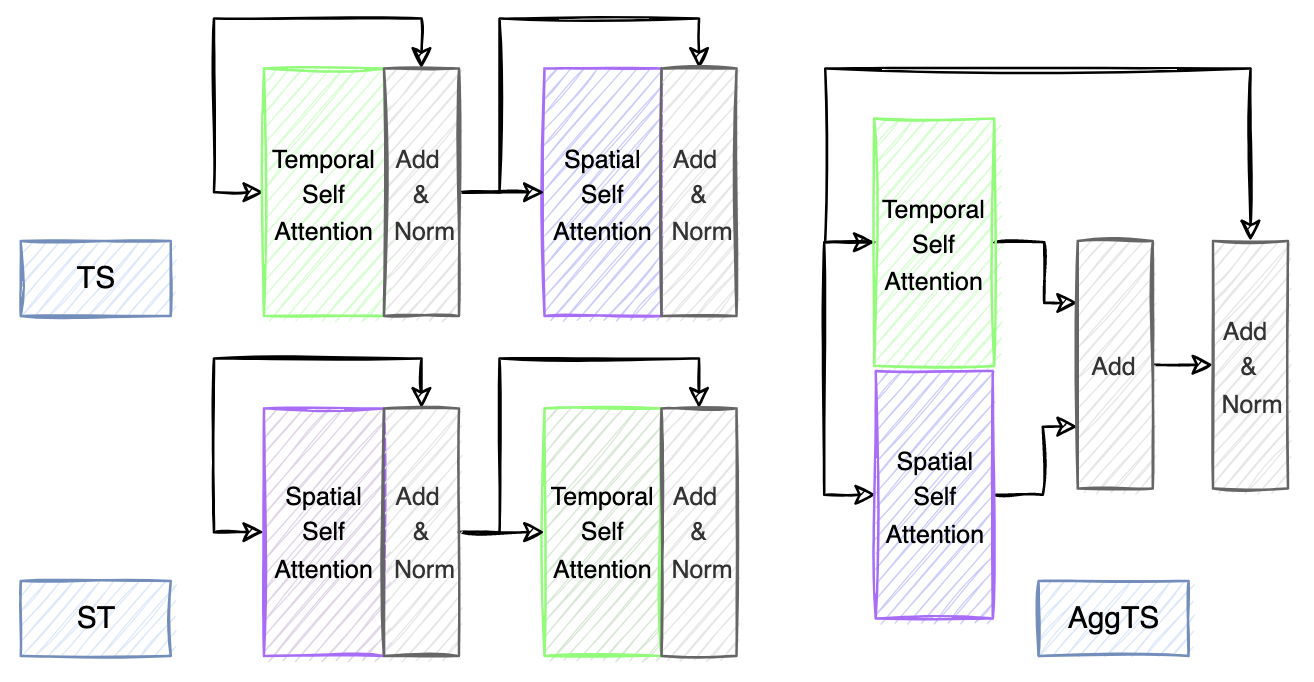}
\centering
\caption{Ablation on Divided Attention types.}
\label{fig:axial_ab}
\end{figure}

\begin{table}[t]
\caption{Comparison between different types of divided attention modules on the ETH/UCY datasets.}
\begin{center}
        \begin{tabular}{c | c | c | c }
          \toprule
            & \multicolumn{3}{c}{Attention Module} \\
            & TS & ST & Agg_TS \\
            \midrule
            \midrule
            ETH & 0.86/1.64 & \textbf{0.82/1.55} & 0.84/1.59 \\
            Hotel & \textbf{0.3/0.56} & 0.31/0.59 & 0.32/0.6 \\
            Univ & 0.62/1.23 & \textbf{0.60/1.25} & 0.62/1.27 \\
            Zara1 & \textbf{0.42/0.89} & 0.44/0.89 & 0.42/0.9 \\
            Zara2 & 0.35/0.74 & \textbf{0.35/0.73} & 0.36/0.76 \\
            \midrule
            Average & 0.51/1.01 & \textbf{0.50/1.00} & 0.51/1.02 \\
          \bottomrule
        \end{tabular}
\end{center}    
    \label{table:ablation}
\end{table}

\subsection{Ablation Studies}

We performed ablation studies by testing three types of factorized attentions (Fig. \ref{fig:axial_ab}):
\begin{enumerate*}
\item TS: the input sequence will go into temporal attention followed by spatial attention.
\item ST: the input sequence will go into spatial attention followed by temporal attention.
\item Agg\_TS: the input sequence will independently go into temporal and spatial attention.
\end{enumerate*}
The resulted sequences will be aggregated into the final sequence. We choose an addition operator for the aggregation. Table \ref{table:ablation} shows that the performances of the TS and ST designs are very close with a slight advantage for the ST module. The Agg\_TS has a relatively lower performance.

\section{Pedestrian trajectory prediction as a non-autoregressive problem}\label{comma}

In previous sections of the paper, we presented the idea of using a parallel decoding strategy to reduce training and inference time alongside the need to eliminate the problem of exposure bias. 
However, the success of this model pushes us to rethink the pedestrian trajectory problem and understand the reason behind this success. In fact, in order to forecast the future states for a given pedestrian, it is more crucial to first understand and predict its latent intention or goal as a whole rather than the details of its trajectory one step at a time. By doing that, it becomes easier to fill in the gaps and generate the rest of the future steps. Indeed, this assumption also holds for multi-agent trajectory prediction by formalizing the key agent-to-agent interaction outcomes. This assumption states that trajectory prediction is based on the information available in the seed sequence and the formulated goals rather than the model's step-by-step target predictions.
Recently, the authors in \cite{ren2020study} conducted a study on non-autoregressive models for sequence generation. They designed an experimental method to estimate how much a given task can be solved or not as a non-autoregressive task. To this end, they designed a self-supervised model, called CoMMA, to quantify the dependence on other targets while predicting a specific target token. The CoMMA model is based on a temporal mixed-attention module introduced in \cite{he2018layer} and is trained using the masked conditional language modeling such as in \cite{devlin2018bert}. In short, a portion of the targets is masked with a probability $p$ and then the missing tokens are reconstructed during the training stage. Afterward, to measure how much a task is non-autoregressive, the token-dependency is estimated using the attention density ratio $R$ for a given masking probability $p$. The probability $p$ ranges uniformly between $(0,1)$, where $p=1$ yields a non-autoregressive model.
The density ratio $R$ represents the ratio in which the model gives attention to the target sequence rather than the seed sequence. In this manner, the higher $R$ is, the more a task is autoregressive.
We define $R(p)$ by:

\begin{equation}
\begin{split}
    R(p) &= \frac{1}{D_{size} } \frac{1}{|M^p|} \sum_{b=1}^{D_{size}} \sum_{i \in M^p} \alpha_i \; ,
\end{split}
\end{equation}

\begin{equation}
\begin{split}
    \alpha_i &= \frac{\frac{1}{T_{pred}} \sum_{j=T_{obs}}^{T_{obs} + T_{pred}} A_{i,j}}{
    \frac{1}{T_{obs}} \sum_{j=1}^{T_{obs}} A_{i,j} + \frac{1}{T_{pred}} \sum_{j=T_{obs}}^{T_{obs} + T_{pred}} A_{i,j}}
\end{split}
\end{equation}

$A_{ij}$ represents the attention from a token $i$ to a token $j$ computed using the first model layer and averaged over all its attention heads. Thus, $\alpha_{ij}$ denotes the attention density ratio when predicting a target $i$. $T_{obs}$ and $T_{pred}$ are the observation and prediction length respectively. In this way, we measure $R(p)$ by averaging $\alpha_i$ for all masked target tokens $M^p$ in a dataset of size $D_{size}$.

In this section, we justify the usage of non-autoregressive parallel decoding for the trajectory prediction task by calculating the attention density ratio \cite{ren2020study} on targets and comparing it to other task types. This justification is complementary to what we have empirically shown in section \ref{results}.

\begin{figure}[t]
  \includegraphics[scale=0.265]{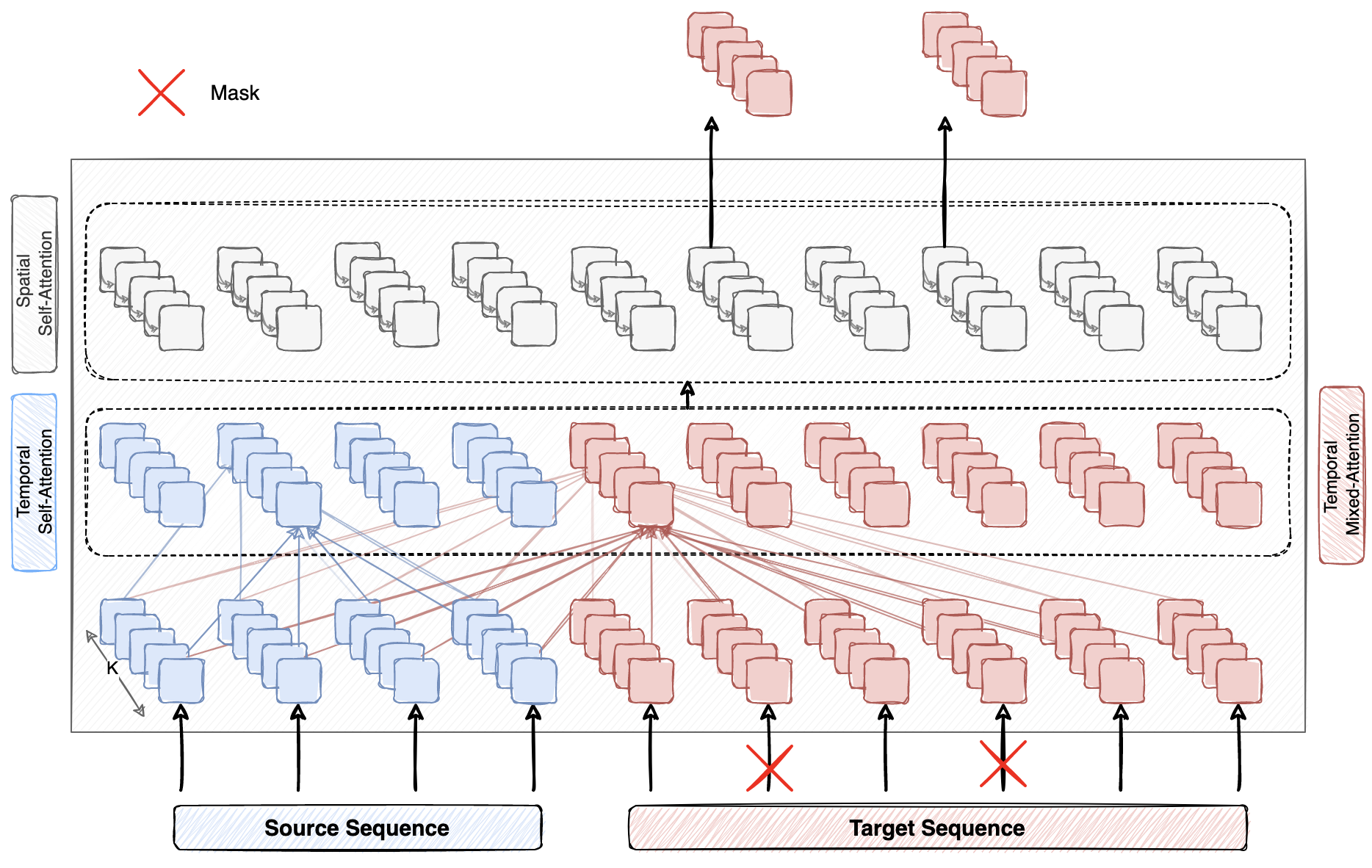}
\centering
\caption{Spatio-temporal Attention module in ST-CoMMA.}
\label{fig:comma}
\end{figure}

\subsection{Model Design}
The CoMMA model is a temporal model that cannot handle multi-agent interactions. It is why we extended its architecture to have a multi-agent spatio-temporal CoMMA model (ST-CoMMA, Fig. \ref{fig:comma}) with a divided attention module similar to what we have introduced in section \ref{axial}. Furthermore, we modified the temporal module to apply a self-attention mechanism on the source sequence and a mix-attention mechanism for target tokens $i$ relative to all sequence tokens $j$. This formalism ensures that the $\sum_j {A_{i,j}}=1$ overall tokens in the sequence. Then, the temporal attention is followed by a spatial self-attention between all agents at each time step. Similar to PreTR, we've added time and agent encoding to the learned embedding module. Additionally, we've used segment embedding to distinguish between source and target tokens as in \cite{devlin2018bert, ren2020study}.

\subsection{Implementation details}
During Bert-like \cite{devlin2018bert} models training, we mask with some probability $p$ a subset of the input tokens using a mask token and train the model to predict these masked tokens. In our experiments, if we choose a time step for masking, then we apply the mask for all agents presented at that time step. 
In general, the model solves a classification problem, where each of the possible input tokens has a class, including the mask token. To follow the same idea, we apply quantization on the scenes by transforming the continuous input into a sequence of word tokens from a vocabulary of size 6827 classes. The quantization step is the average difference in distance between two consecutive steps for pedestrians in all training scenes. We trained the model with $2$ layers and $8$ attention heads. The embedding dimension $d$ is set to $512$ and the $Pff$ layer hidden dimension to $1024$. 

\subsection{Attention density ratio results}
\begin{figure}[t]
\begin{tikzpicture}[scale=0.9, transform shape]
\begin{axis}[
    xlabel={$p$},
    ylabel={$R(p)$},
    xmin=0.08, xmax=0.52,
    ymin=0.4, ymax=0.7,
    xtick={0.1,0.2,0.3,0.4,0.5},
    ytick={0.4,0.45,0.5,0.55,0.6,0.65},
    legend style={at={(0.03,0.46)},anchor=west},
    ymajorgrids=true,
    grid style=dashed,
]

\addplot[
    color=red,
    mark=square,
    mark size=3.5pt,
    ]
    coordinates {
    (0.1,0.4335)(0.2,0.4326)(0.3,0.4317)(0.4,0.4306)(0.5,0.4277)
    };

\addplot[
    color=green,
    mark=o,
    mark size=3.5pt,
    ]
    coordinates {
    (0.1,0.48)(0.2,0.475)(0.3,0.471)(0.4,0.47)(0.5,0.465)
    };

\addplot[
    color=blue,
    mark=triangle,
    mark size=3.9pt,
    ]
    coordinates {
    (0.1,0.64)(0.2,0.62)(0.3,0.608)(0.4,0.595)(0.5,0.591)
    };
    
\addplot[
    color=orange,
    mark=*,
    mark size=3.5pt,
    ]
    coordinates {
    (0.1,0.68)(0.2,0.675)(0.3,0.672)(0.4,0.671)(0.5,0.669)
    };
    
    \legend{Trajectory, TTS, NMT, ASR};
    
\end{axis}
\end{tikzpicture}
\caption{Attention density ratio $R(p)$ on target tokens with a masking probability $p \in (0.1,0.5)$.}
\label{fig:rp}
\end{figure}
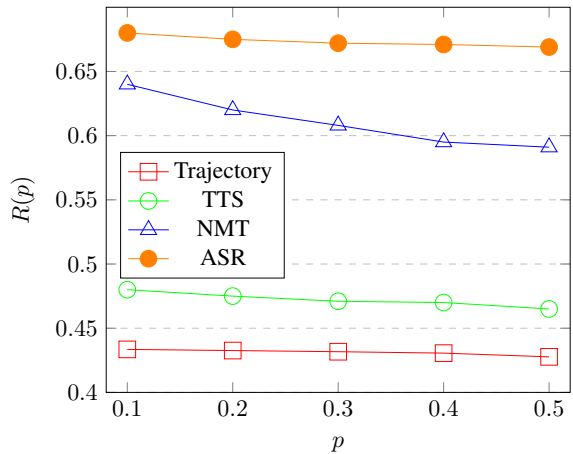

After training the model, we calculated $R(p)$ for different values of $p$ varying from $0.1$ to $0.5$. We compared the $R(p)$ score on the trajectory prediction task with the scores estimated in \cite{ren2020study} for the following language modeling tasks: neural machine translation (NMT), automatic speech recognition (ASR), and text-to-speech (TTS). As we can see in (Fig. \ref{fig:rp}) the trajectory prediction task has the lowest $R(p)$ score for all $p$ values. This result suggests clearly that the trajectory prediction task tends to be non-autoregressive by nature. In fact, to predict a target token, the model pays more attention to the source sequence than to other targets, even in their presence. The findings of this experiment went beyond the results in \cite{ren2020study}, showing that a model with a low $R$ score can not only shrink the performance gap between autoregressive and non-autoregressive predictions but can also have an advantage by achieving even better results. In fact, trajectory prediction planning-based methods \cite{ziebart2008maximum, kitani2012activity} assume that a pedestrian is a rational agent in which we can formulate its latent intents. The interpretation behind our results is similar. The pedestrian motion can be stochastic in terms of multi-modality but, in principle, rational rather than chaotic \cite{kincanon1999chaos}. However, it will be hard to predict the future trajectory of a non-rational agent without conditioning on the target's previous predictions.

\section{Conclusion}
In this paper, we tackled the problem of multi-agent trajectory prediction by proposing a Transformer based interaction-aware model. We showed that a factorized version of the spatio-temporal attention module is best suited for trajectory prediction by achieving better performances and faster computational time than previous works. In addition, we introduced a parallel decoding strategy to our architecture that can outperform autoregressive decoding and show less computational complexity with a $13 \times$ speed-up on the $4.8s$ prediction horizon. Finally, we raised the question about the nature of the trajectory prediction task, whether more fitted for autoregressive or parallel prediction mode. We extended a self-supervised model to cover spatio-temporal features and showed that the trajectory prediction task has a non-autoregressive behavior and is suitable for parallelization. Based on these findings, we suggest shifting the research interest in trajectory prediction toward the non-autoregressive models to enhance this promising research area. 

\section*{Acknowledgement}
This work was carried out in the framework of the OpenLab ``Artificial Intelligence'' in the context of a partnership between INRIA institute and Stellantis company.

\end{document}